\documentclass[lettersize,journal,10pt]{IEEEtran}
\usepackage{amsmath,amsfonts}
\usepackage{algorithm}
\usepackage{algpseudocode}
\usepackage{array}
\usepackage[caption=false,font=normalsize,labelfont=sf,textfont=sf]{subfig}
\usepackage{textcomp}
\usepackage{stfloats}
\usepackage{url}
\usepackage{verbatim}
\usepackage{graphicx}
\usepackage{cite}
\usepackage{tabularx}
\usepackage{multirow}
\usepackage{adjustbox}
\usepackage{authblk}

\newcolumntype{Y}{>{\centering\arraybackslash}X}

\begin{document}

\title{DECADE: A Temporally-Consistent Unsupervised Diffusion Model for Enhanced \textsuperscript{82}Rb Dynamic Cardiac PET Image Denoising}

\author[a]{Yinchi Zhou}
\author[a]{Liang Guo}
\author[a]{Huidong Xie}
\author[a]{Yuexi Du}
\author[b]{Ashley Wang}
\author[a]{Menghua Xia}
\author[a]{Tian Yu}
\author[c]{Ramesh Fazzone-Chettiar}
\author[c]{Christopher Weyman} 
\author[d]{Bruce Spottiswoode}
\author[d]{Vladimir Panin}
\author[e]{Kuangyu Shi}
\author[a,b,c]{Edward J. Miller} % cardiology 
\author[c]{Attila Feher} % cardiology 
\author[a,b,c]{Albert J. Sinusas}
\author[a,b]{Nicha C. Dvornek}
\author[a,b]{Chi Liu}
\affil[a]{Department of Biomedical Engineering, Yale University, New Haven, CT, USA}
\affil[b]{Department of Radiology and Biomedical Imaging, Yale School of Medicine, New Haven, CT, USA}
\affil[c]{Department of Medicine (Cardiology), Yale School of Medicine, New Haven, CT, USA}
\affil[d]{Siemens Medical Solutions USA, Inc., Knoxville, TN, USA}
\affil[e]{Department of Nuclear Medicine, Inselspital, Bern University Hospital, University of Bern, Bern, Switzerland}

% \author{Yinchi Zhou,~\IEEEmembership{Staff,~IEEE,}}
        % <-this % stops a space



% \begin{figure*}
%     \centering
%     \includegraphics[width=\linewidth]
%     {figure/graphic_abstract.png}
%     \caption{graphic abstract}
%     \label{framework}
% \end{figure*}

\newpage

\maketitle

\begin{abstract}
\textsuperscript{82}Rb dynamic cardiac PET imaging is highly effective for the clinical diagnosis of coronary artery disease (CAD). However, its short half-life introduces significant noise, compromising the image quality of dynamic frames and parametric imaging. The general absence of paired clean-noisy dynamic frames for \textsuperscript{82}Rb PET, along with the rapid changes in tracer distribution and varying noise levels across dynamic frames, pose challenges for the application of current deep learning denoising approaches. We developed DECADE (A Temporally-Consistent Unsupervised \textit{D}iffusion model for \textit{E}nhanced \textsuperscript{82}Rb \textit{CA}rdiac PET \textit{DE}noising), a novel unsupervised diffusion model framework capable of generalizing across dynamic frames from early to late phase. DECADE ensures quantitative and qualitative accuracy by incorporating temporal consistency into both diffusion training and the sampling stage, using noisy frames as image guidance in its iterative sampling process. The framework was trained and evaluated on datasets acquired from Siemens Vision 450 and Siemens Biograph Vision Quadra scanners.  For the Vision 450 dataset, DECADE consistently produced high-quality dynamic and parametric images with reduced noise while preserving myocardial blood flow (MBF) and myocardial flow reserve (MFR) quantification. For the Quadra dataset, using 15\%-count as noisy input and full-count as the ground truth, DECADE demonstrated superior performance in image quality and consistent $K_1$ and MBF quantification compared to other UNet-based models and diffusion models. DECADE effectively generated denoised $K_1$ maps from 15\%-count images that are consistent with those derived from full-count images. The proposed DECADE framework enables effective unsupervised denoising of \textsuperscript{82}Rb dynamic cardiac PET without the need for paired training data. This approach facilitates clearer visualization of dynamic frames and parametric images while preserving quantitative integrity, demonstrating strong potential for clinical application in \textsuperscript{82}Rb dynamic cardiac PET imaging denoising.

\end{abstract}


\setcounter{figure}{0}

\section{Introduction}

% introduce cardiac pet, dynamic cardiac pet with rb-82
Positron emission tomography (PET) myocardial perfusion imaging provides a non-invasive assessment to detect coronary artery disease (CAD) with high accuracy \cite{prior2012quantification, schindler2010cardiac, carli2006integrated}. Following the administration of a radioactive tracer, dynamic frame sequences are typically acquired over a few minutes, depending on the tracer type, until the myocardium is sufficiently perfused. The absolute quantification of myocardial blood flow (MBF) and myocardial flow reserve (MFR) can then be derived by tracking dynamic tracer activities in the blood and myocardium over time in dynamic cardiac PET. Specifically, 3D parametric images of pharmacokinetic parameters are reconstructed from 4D cardiac dynamic scans using kinetic modeling to quantify blood flow. Quantification of MBF and MFR has demonstrated enhanced diagnostic and prognostic effectiveness \cite{schindler2010cardiac, sciagra2021eanm, neglia2009myocardial}. Among the various PET tracers, Rubidium-82 (\textsuperscript{82}Rb) is notable for its short half-life (76.4 seconds), which enables rapid sequential stress-rest imaging within a clinically manageable time frame (30-40 minutes). It also  provides relatively good image quality and overall sensitivity for diagnosing CAD. The short half-life of \textsuperscript{82}Rb also results in a lower effective radiation dose, halving the radiation exposure for patients compared to \textsuperscript{99m}Tc-based radiotracers \cite{chatal2015story}. Moreover, \textsuperscript{82}Rb is a generator product and does not require an onsite cyclotron, making it the most widely used PET tracer for myocardial perfusion imaging in the USA \cite{driessen2017myocardial}.

% challenges with rb-82
Despite these advantages, \textsuperscript{82}Rb PET imaging presents several unique challenges. Due to its short half-life, individual dynamic frames, especially late frames, can exhibit substantial noise over time, and the time-activity curves derived from \textsuperscript{82}Rb images can be inconsistent. This results in degraded image quality of reconstructed parametric images and inaccurate voxel-level MBF and MFR quantification \cite{ghotbi2014comparison, mohy2015quantitative}. Therefore, the development of effective denoising techniques for \textsuperscript{82}Rb dynamic cardiac PET is imperative. Traditional methods such as image smoothing \cite{ghotbi2014comparison, chilra2017cardiac} and wavelet transforms \cite{lin2001improving} have been applied to suppress noise in \textsuperscript{82}Rb PET images. Nevertheless, these methods often lead to information loss due to excessive blurring, adversely affecting the quantitative accuracy of kinetic parameters \cite{xu2018joint, le2010enhanced, mohy2015quantitative, wang2020pet}.

% machine learning methods for denoising
In recent years, deep learning algorithms have outperformed conventional methods for PET image denoising \cite{cui2019pet, song2021noise2void, zhou2020supervised, xia2025leqmod, xue2022cross}. Various model architectures have been explored. Xu et al. employed a UNet \cite{ronneberger2015u} for reconstructing ultra-low-dose PET images (0.5\% of the normal dose) through residual learning \cite{xu2017200x}. Wang et al. developed an adversarial training scheme using conditional GANs to recover full-dose PET images from low-dose PET and further enhanced the performance by incorporating MRI images \cite{wang20183d, wang20183d2}. Jang et al. implemented a spatial and channel-wise encoder-decoder transformer to leverage local and global information, enhancing denoising performance compared to CNN-based models \cite{jang2023spach}. More recently, denoising diffusion probabilistic models (DDPMs) \cite{ho2020denoising} have become state-of-the-art generative models, demonstrating superior performance in image translation, image generation, and image restoration tasks \cite{li2023bbdm, zhou2024cascaded, saharia2022palette, dhariwal2021diffusion, wu2024medsegdiff, bansal2023cold}. Diffusion models have also been successfully applied to PET denoising \cite{gong2024pet, xie2024dose, yu2024pet, zhou2025fed, chen20252}.

% unsupervised learning, over smooth images or denoising quality is limited
Nonetheless, the direct application of these methods to \textsuperscript{82}Rb dynamic cardiac PET denoising encounters significant challenges. Most algorithms rely on supervised learning, requiring paired low-dose and full-dose images for training, which are challenging to obtain for \textsuperscript{82}Rb dynamic PET due to its short half-life, unless an ultra high sensitivity scanner is used (e.g. Long Axial Field of View (LAFOV) PET). Moreover, individual frames contain substantial noise, and high-quality counterparts for each frame are unavailable. Unsupervised learning approaches, such as deep image prior \cite{ulyanov2018deep}, have been investigated for PET denoising. Deep image prior utilizes only noisy PET frames during training and a static image as the network input, iteratively generating denoised dynamic PET images. Self-supervised convolutional neural networks (CNN) like Noise2Void \cite{song2021noise2void} have also demonstrated potential for PET denoising. For \textsuperscript{82}Rb dynamic PET, considering the varying noise levels across frames, Xie et al. encoded noise level information into the self-supervised network to enhance denoising \cite{xie2025noise}. However, deep image prior requires re-training for each subject, complicating integration into clinical workflows. Additionally, CNN-based methods tend to over-smooth images and underestimate biological parameters \cite{xie2025noise, xia2025anatomically}, and they fail to incorporate temporal information for dynamic PET, treating each frame independently.

\begin{figure*}[htb!]
    \centering
    \includegraphics[width=\linewidth]{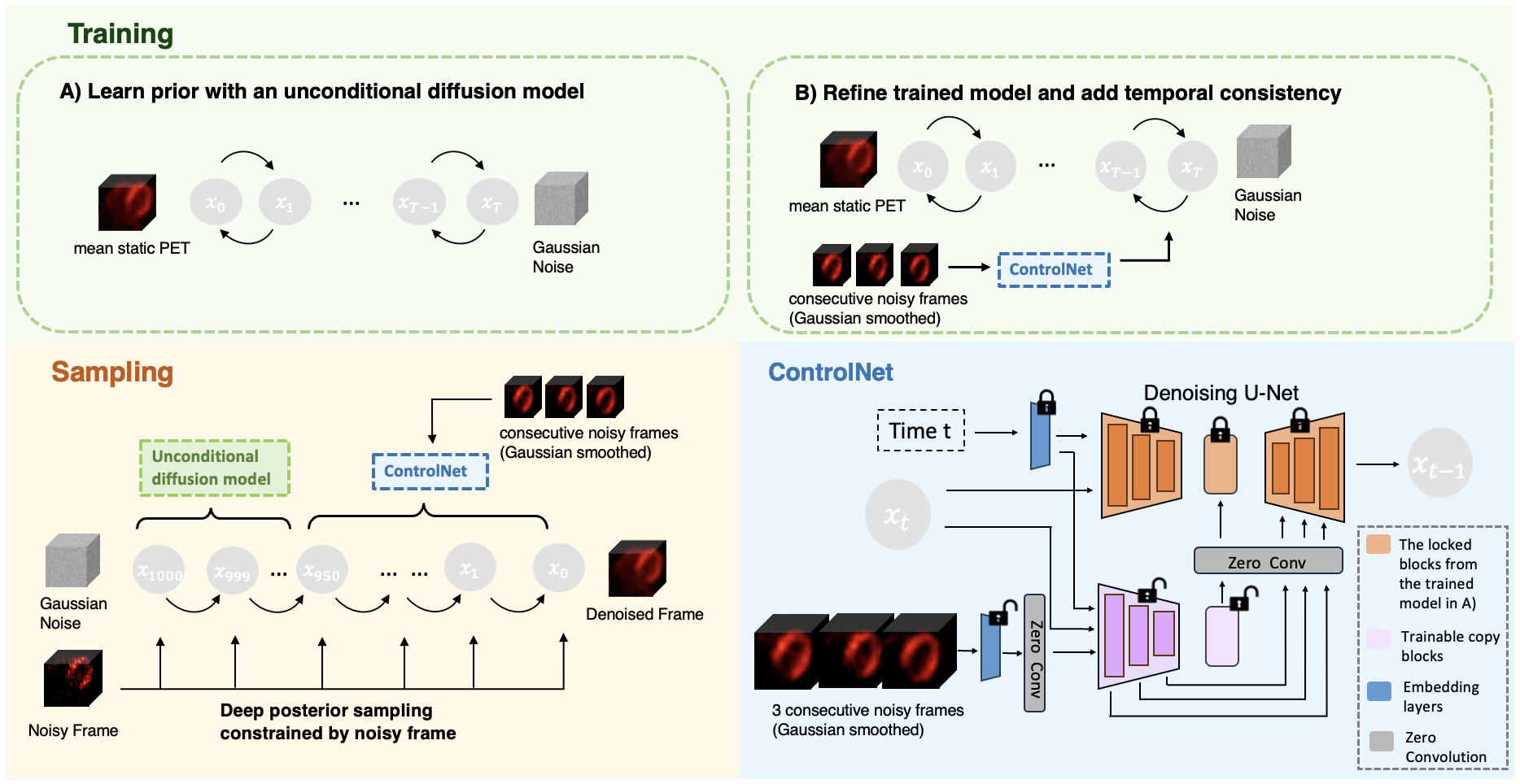}
    \caption{DECADE consists of a diffusion model training stage and a sampling stage. During Training Stage A, the model is pre-trained on 2-7 minute mean static PET images. Three Gaussian-smoothed consecutive noisy frames are then used as conditional input into ControlNet to refine the pre-trained model. In the sampling stage, the unconditional diffusion model is applied for time steps from 1000 to 950, followed by the use of ControlNet from time steps 950 to 0. Noisy frames are incorporated at each time step to guide the denoising process.}
    \label{framework}
\end{figure*}

% proposed methods 
To address these challenges, we propose DECADE, a temporally-consistent unsupervised \textbf{\textit{D}}iffusion model for \textbf{\textit{E}}nhanced \textsuperscript{82}Rb \textbf{\textit{CA}}rdiac PET \textbf{\textit{DE}}noising. DECADE enhances denoising performance by leveraging consecutive frames to ensure temporal consistency and utilizing individual noisy frames to guide the denoising process. Specifically, since static images are commonly used in myocardial perfusion imaging to identify regional abnormalities and are widely available, our unsupervised learning approach involves pre-training a diffusion model on mean static images with reduced noise to capture the clean data distribution. This is followed by fine-tuning the pre-trained model with Gaussian-smoothed consecutive noisy frames through ControlNet \cite{zhang2023adding} to maintain temporal consistency. Additionally, we introduce a novel sampling strategy that iteratively incorporates individual noisy frames at each diffusion time step to guide the denoising trajectory. The strategy exploits a fine-tuned model trained with temporally consistent consecutive frames, using these temporal features to enhance structural detail recovery for improved visual fidelity, while the integration of noisy frames provides a quantitative anchor that preserves measurement fidelity. We validated our proposed method using datasets acquired from Siemens Vision 450 and Siemens Quadra scanners. Our contributions can be summarized as follows: 1) We developed an unsupervised learning framework with diffusion models for dynamic cardiac PET image denoising, generalizing to both early and late frames. 2) We incorporated temporal information from dynamic frames, ensuring temporal consistency during the diffusion model training and sampling stages to enhance denoising. 3) We proposed a novel diffusion sampling process that integrates image guidance and temporal consistency for image denoising. 4) We validated our methods using high count LAFOV PET data, demonstrating state-of-the-art performance in improving dynamic frame image quality and reducing noise in parametric images. Our method successfully recovered the image quality of noisy dynamic frames of \textsuperscript{82}Rb cardiac PET, while preserving anatomical details and MBF and MFR quantification accuracy.

\section{Materials and Methods}
\subsection{Data Acquisition and Pre-processing}
\textbf{Vision 450 Dataset}: A total of 384 anonymized patients that underwent rest/stress \textsuperscript{82}Rb imaging on a Siemens Vision 450 PET/CT system at Yale PET Center were included in this study, with the approval of the Yale Institutional Review Board. The cohort consisted of 186 female and 198 male patients. The dataset was split into 329 subjects for training, 5 for validation, and 50 for testing. The average body mass index (BMI) across all patients was 34.02 $\pm$ 9.67 across all patients. Among the 384 patients, 37 were referred to a follow-up catheterization procedure. The average injection dose across all patients was 826.78 $\pm$ 127.07 MBq. For each scan, the 7-minute list-mode data were re-binned into 30 dynamic frames (14 x 5 s, 7 x 10 s, 4 x 20 s, 4 x 30 s, 1 x 80 s). Mean static PET images were reconstructed using list-mode data from 2-7 minutes. All frames were reconstructed using the Ordered Subsets Expectation Maximization (OSEM) algorithm with 2 iterations and 5 subsets. A 2 mm FWHM Gaussian filter was applied post-reconstruction. The reconstructed images have dimensions of 440 x 400 x 119 with a voxel size of 1.65 x 1.65 x 1.65 mm\textsuperscript{3}. The cardiac regions were cropped from the reconstructed images using heart labels obtained from TotalSegmentator on CT images, with a dimension of 120 x 100 x 96 \cite{wasserthal2023totalsegmentator}. Clinical reports for each patient were available to validate the results.

\textbf{Quadra Dataset}: This dataset was collected on a Siemens Quadra PET/CT system at the Department of Nuclear Medicine, University of Bern, Switzerland. It comprises five \textsuperscript{82}Rb subjects with repeated rest and stress scans, including 2 female and 3 male patient. The average BMI across all patients was 29.55 $\pm$ 6.63, and the average injection dose is 429.20 $\pm$ 49.64 MBq. For each scan, the 7-minute list-mode data were re-binned into 30 dynamic frames (14 x 5 s, 7 x 10 s, 4 x 20 s, 4 x 30 s, 1 x 80 s). Given that Quadra scanners have higher sensitivity, allowing for higher image quality, we downsampled the list-mode data from Quadra scans to images with similar noise levels acquired on Vision 450 scanners, using full-count images as the gold standard for validation with the denoised results of downsampled data. The sensitivity of Quadra scanner is approximately 6.7 times that of the Vision 450, so each dynamic frame of Quadra scans was randomly downsampled to a 15\% count level to simulate low-count noisy frames while using the Quadra scans as full-count references \cite{prenosil2022performance, carlier2020pmt}. Both 15\% count and full-count scans were reconstructed using the OSEM algorithm with 2 iterations and 21 subsets. A 2 mm FWHM Gaussian filter was applied post-reconstruction. The reconstructed images initially had dimensions of 220 x 220 x 644 with a voxel size of 3.3 x 3.3 x 1.65 mm\textsuperscript{3}. The reconstructed images were subsequently resampled to a voxel size of 1.65 x 1.65 x 1.65 mm\textsuperscript{3}, resulting in a final matrix size of 440 x 440 x 644. The cardiac regions with dimensions of 120 x 100 x 96 were manually cropped from the whole-body images.

\subsection{Framework Overview}
Our proposed framework consists of a two-stage training module and a sampling module (Fig.~\ref{framework}). During Phase A of the training stage, we leverage widely available low-noise static cardiac PET images for pre-training to learn the clean data distribution. In Phase B, dynamic PET data are incorporated into the fine-tuning process through ControlNet \cite{zhang2023adding}, integrating spatial control conditions into the pre-trained diffusion models to enhance performance in dynamic frame denoising. Specifically, three consecutive Gaussian-smoothed dynamic frames are used as control inputs to guide the denoising process and ensure temporal consistency. This integration addresses artifacts that may arise due to discrepancies between the pre-training data from mean static PET and the target distribution of clean dynamic images, especially in regions with low uptake, such as myocardial defect regions, anatomical boundaries, or ventricular cavities, where hallucinations or artifacts might occur.

For the sampling process, we propose a diffusion sampling strategy based on deep posterior sampling (DPS) \cite{ulyanov2018deep} to integrate the original single noisy frames as quantitative constraints when sampling from the learned data distribution, ensuring image domain consistency. Simultaneously, Gaussian-smoothed consecutive frames are incorporated to ensure temporal consistency and further refine the structural details of denoised results in the intermediate steps. During the backward process, the trained unconditional diffusion model is used exclusively in early time steps to allow for the appropriate approximation of clean images given noisy inputs. From intermediate time steps onward, when major coarse, low-frequency components have been recovered, the addition of ControlNet introduces temporal consistency, further refining the generated images until the denoised images are obtained after 1000 sampling steps.

\subsection{3D Denoising Diffusion Probabilistic Model}
Diffusion models are generative models defined through a Markov chain over latent variables $x_1, x_2, ..., x_T$ \cite{ho2020denoising}. They include a forward and a reverse process governed by a Markov chain. In the forward process, small Gaussian noise $\epsilon\sim\mathcal{N}(0,I)$ is gradually added to the target image $x_0 \sim q(x_0)$ over $T$ time steps. The Markov property allows us to sample any step conditioned on $x_0$. The noisy sample $x_t$ at time step $t$ is defined as follows: 
\begin{equation}
    q(x_t|x0) = \mathcal{N}(x_t; \sqrt{\bar{\alpha}_t}x_0, (1-\bar{\alpha}_t)\mathbf{I})
\end{equation}
\begin{equation}
\label{forward}
    x_t = \sqrt{\bar{\alpha_t}}x_0 + (1-\bar{\alpha_t})\epsilon
\end{equation}
where $\beta$ is the variance schedule, $\alpha = 1 - \beta$\, and $\bar{\alpha} = \prod_{s=0}^t\alpha_s$. With a sufficiently large $T$ and appropriate variance scheduling, $x_T$ becomes a nearly isotropic Gaussian distribution. In the reverse process, the goal is to infer $x_{t-1}$ given $x_t$. Based on Bayes theorem, the posterior $q(x_{t-1}|x_t)$ follows a Gaussian distribution and is tractable when conditioned on $x_0$:
\begin{equation}
\label{qx}
    q(x_{t-1} | x_t, x_0) = \mathcal{N}(x_{t-1}; \mu_t(x_t,x_0), \frac{1-\bar{\alpha}_{t-1}}{1-\bar{\alpha_t}}\beta_t\mathbf{I})
\end{equation}
\begin{equation}
\label{mu0}
    \text{where } \mu_t(x_t,x_0) = \frac{\sqrt{\bar{\alpha}_{t-1}}\beta_t}{1-\bar{\alpha_t}}x_0 + \frac{\sqrt{\alpha_t}(1-\bar{\alpha}_{t-1})}{1-\bar{\alpha_t}}x_t
\end{equation}

Instead of approximating $\mu_\theta$ with a neural network, Ho et al. \cite{ho2020denoising} proposed to approximate the noise $\epsilon$ in Eq.~\eqref{forward} by a trained network $\epsilon_\theta (x_t, t)$ with parameters $\theta$, which could also be interpreted as the gradient of the data log-likelihood, known as the score function. Starting with  $x_T$, a clean sample $x_0$ can be iteratively obtained. Given $x_t$, the reverse process can be written and parameterized with $\epsilon_{\theta}$ using Eq.~\eqref{forward} and Eq.~\eqref{qx}: 

\begin{align}
   p_\theta(x_{t-1}|x_t) = & \mathcal{N}(x_{t-1};\mu_\theta(x_t,t),{\sigma_t}^2\mathbf{I}) \\
   \text{where } \mu_\theta(x_t,t) = &\frac{1}{\sqrt{\alpha_t}}(x_t-\frac{1-\alpha_t}{\sqrt{1-\bar{\alpha}_t}}\epsilon_\theta(x_t,t))
\end{align}
The training objective is then to minimize:

\begin{equation}
    \mathbb{E}_{t,x_t,\epsilon}[{||\epsilon-\epsilon_\theta(x_t,t)||}^2]
\end{equation}

\subsection{Image Guidance for Diffusion Sampling}
PET denoising is intrinsically an inverse problem where the goal is to retrieve the clean image $x$ given the noisy image $y$, which is known as posterior sampling $p(x|y)$. We adapted the deep posterior sampling (DPS) for diffusion models proposed by Chang et. al \cite{chung2022diffusion} to incorporate the noisy image $y$ into the diffusion sampling process, adding image guidance from the noisy frames to help preserve structure and quantification. In DDPM sampling, $p(x_0|x_t)$ has a posterior mean $\hat{x_0}$ approximated with the score function $\epsilon_\theta(x_t, t)$:

\begin{equation}
\label{x0}
    \hat{x_0} \simeq \frac{1}{\sqrt{\bar{\alpha_t}}} (x_t + (1-\bar{\alpha_t})\epsilon_\theta (x_t, t))
\end{equation}
Based on Eq.~\eqref{mu0} and Eq.~\eqref{x0}, plugging in $\hat{x_0}$,

\begin{equation}
    x_{t-1} = \frac{\sqrt{\bar{\alpha}_{t-1}}\beta_t}{1-\bar{\alpha_t}}\hat{x_0} + \frac{\sqrt{\alpha_t}(1-\bar{\alpha}_{t-1})}{1-\bar{\alpha_t}}x_t + \sigma_t\textbf{z}
\end{equation}
where $z\sim N(0, \mathbf{I})$ and $\sigma_t$ is the fixed constant value based on $\beta_t$.

For DPS, the measurement model $p(y|x_t)$ can be approximated with $p(y|x_0)$:
\begin{equation}
\label{yx0}
    p(y|x_t) \simeq p(y|\hat{x_0})
\end{equation}
For Gaussian noise, given the noisy image $y$ and clean image $x$, the likelihood function is
\begin{equation}
    p(y|x_0) = \frac{1}{\sqrt{(2\pi)^n\sigma^{2n}}}\text{exp}[-\frac{||y-x_0||^2_2}{2\sigma^2}]
\end{equation}
where $n$ denotes the dimension of the measurement $y$. By differentiating $p(y|x_t)$ with respect to $x_t$ using Eq.~\eqref{yx0},
\begin{align}
    \nabla_{x_t} \text{log }p(y|x_t) \simeq -\frac{1}{\sigma^2}\nabla_{x_t}||y-\hat{x_0}(x_t)||^2_2 \\
    \nabla_{x_t} \text{log }p(x_t|y) \simeq \epsilon_\theta(x_t,t) - \rho\nabla_{x_t}||y-\hat{x_0}(x_t)||^2_2 
\end{align}
where $\rho$ is the step size for the regularization at each time step $t$. With $\hat{x_0}$ updated at each time step $t$, the final $x_{t-1}$ can be obtained as 

\begin{equation}
\label{xt-1}
\begin{aligned}
    x_{t-1} &= \frac{\sqrt{\bar{\alpha}_{t-1}}\beta_t}{1-\bar{\alpha_t}}\hat{x_0}(x_t) + \frac{\sqrt{\alpha_t}(1-\bar{\alpha}_{t-1})}{1-\bar{\alpha_t}}x_t \\
    & + \sigma_t\textbf{z} - \rho\nabla_{x_t}||y-\hat{x_0}(x_t)||^2_2
\end{aligned}
\end{equation}

For dynamic PET denoising, phase A trains an unconditional diffusion model by taking advantage of abundant mean static images. To better align the denoised images and noisy frames, image consistency is enforced during the sampling process using DPS to guide the generation process, allowing unsupervised learning for dynamic PET denoising. An exponential decay scheduler is used for the step size $\rho$, with the constraint contributing more in the early sampling stage to guide the diffusion models to recover low-frequency information. As the rough shape of the images is recovered, the diffusion prior plays a more important role in the later sampling stages. In this study, when sampling 1000 time steps, the step size for measurement constraints is set to $\textit{exp}^{-0.05(1000-t)}\frac{w}{||y-\hat{x_0}(x_t)||^2_2}$ where $w$ is a fixed hyperparameter to determine the constraint weight.

\subsection{Temporal-Consistency for Diffusion Model Training and Sampling}

While the two-stage training scheme and image guidance in sampling ensure image consistency and preserve quantification, temporal information is still not considered. Therefore, neighboring frames are integrated into Phase B through ControlNet to provide contextual and temporal information from consecutive frames. This integration helps increase the robustness of model in regions that are challenging to denoise, such as low-uptake regions where artifacts might occur. The underlying assumption is that important anatomical structures, such as the ventricles, aorta, or myocardium, remain consistent across dynamic frames. Contextual information from neighboring frames  ensures the consistency of these anatomical structures over time. DECADE integrates temporal consistency into both the training and sampling processes by using Gaussian-smoothed consecutive frames to inform the model of the temporal context, thereby enhancing its ability to recognize and preserve anatomical consistency. This results in more accurate and reliable denoising across all dynamic frames, mitigating potential artifacts.

In Phase B, ControlNet freezes the pre-trained diffusion model and creates a trainable copy that learns from conditional inputs, with a zero convolution on the trainable copy ensuring minimal distortion of the trained diffusion model. For dynamic PET denoising, we trained a ControlNet where the diffusion models are conditioned on three consecutive Gaussian-smoothed noisy frames, adding temporal information to help the model learn the representation of dynamic frames. Specifically, as shown in Fig.~\ref{framework}, consecutive frames are passed through 3-layer 3D convolutional embedding layers to convert the input into latent space feature vectors that match the size of the pre-trained diffusion model, followed by a zero convolution layer. The trainable copy of encoder blocks and middle blocks were created and connected to the pre-trained diffusion U-Net through zero convolution. During training, the parameters of the pre-trained U-Net were frozen, and only the trainable copy blocks and embedding layers for condition frames were fine-tuned. The details of the ControlNet architecture can be found in \cite{zhang2023adding}. 

In the reverse sampling process, the intermediate step $t_c=950$ is empirically selected to switch the model. From $t=950$ to $t=0$ ControlNet is used to refine the high-level features of denoised images. The unconditional diffusion model is used in the early stage $t=1000$ to $t=950$ to facilitate the recovery of low-frequency information using DPS. After initial convergence to relatively aligned results, ControlNet further refines the details. Therefore, from $t=950$ to $t=0$, the conditional temporal frames $c$ are used. In Eq.~\eqref{x0} $\hat{x_0}$ is modified to 
\begin{equation}
    \hat{x_0} \simeq \frac{1}{\sqrt{\bar{\alpha_t}}} (x_t + (1-\bar{\alpha_t})\epsilon_\theta (x_t, t, c))
\end{equation}
and is then plugged into Eq.~\eqref{xt-1} to obtain $x_{t-1}$ at each time step $t$ iteratively until the denoised image $x_0$ is obtained. Our proposed diffusion sampling algorithm is shown as Alg.~\ref{sample_alg}, combining both image guidance and temporal consistency in the sampling process to denoise noisy frames.

\begin{algorithm}[t]
\caption{DECADE Sampling Algorithm}
\label{sample_alg}
\begin{algorithmic} 
\State \textbf{Require}: noisy frame $y$, temporal frames conditions $c$, diffusion steps $T=1000$, constraint weight $w$, intermediate step $t_c$ 
\State $x_T \sim \mathcal{N}(0,\mathbf{I)}$ \Comment{Sample from Gaussian Distribution}
\For{$t=T,T-1,..,1$}
    \If{$t > t_c$}
        \State $\epsilon = \epsilon_\theta(x_t, t)$ \Comment{Use unconditional diffusion model}
    \Else{ $0 < t < t_c$}
        \State $\epsilon = \epsilon_\theta(x_t, t, c)$ \Comment{Condition on temporal frames}
    \EndIf
    \State $\hat{x_0} = \frac{1}{\sqrt{\bar{\alpha_t}}}(x_t + (1-\bar{\alpha}_t)\hat{\epsilon})$ \Comment Approximate $x_0$  \vspace{6pt}
    \State $z \sim \mathcal{N}(0,\mathbf{I)}$ \Comment{Sample added noise} \vspace{6pt}
    \State $x_{t-1} = \frac{\sqrt{\bar{\alpha}_{t-1}}\beta_t}{1-\bar{\alpha_t}}\hat{x}_0(x_t) + \frac{\sqrt{\alpha_t}(1-\bar{\alpha}_{t-1})}{1-\bar{\alpha_t}}x_t + \sigma_t\textbf{z} $ \\ \vspace{2pt} \hfill
    \Comment{Forward diffusion} \vspace{6pt}
    \State $x_{t-1} = x_{t-1} - e^{-0.05(1000-t)}\frac{w}{||y-\hat{x}_0||}^2_2\nabla_{x_t}||y-\hat{x}_0(x_t)||^2_2$   \\  \hfill 
    \Comment{Constraint on the noisy frame}

\EndFor
\State \textbf{return} $\hat{x}_0$

\end{algorithmic}
\end{algorithm}

\subsection{Tracer Kinetic Modeling and Parametric Imaging}
In this study, a three-parameter one-tissue compartment model was employed to analyze dynamic \textsuperscript{82}Rb PET imaging data for the quantification of MBF. This model considers the exchange of the tracer between plasma and myocardial tissue, as well as the contribution of the blood volume. The tissue concentration $C_T$ follows the equation
\begin{equation}
\label{kinetic}
    C_T(t) = K_1C_b(t)\otimes e^{-k_2t} + V_bC_b(t)
\end{equation}
where $C_b(t)$ represents the image-derived input function (IDIF) from the left ventrical, $V_b$ is the fractional blood volume, $K_1$ is the uptake rate, and $k_2$ is the clearance rate. Eq.~\ref{kinetic} was fit to each voxel using the basis function method to generate voxel-wise parametric images \cite{lodge2000parametric}. The generalized Renkin–Crone model was utilized to quantify MBF \cite{renkin1959transport, crone1963permeability}:
\begin{equation*}
    K_1 = MBF(1-ae^{-b/MBF})
\end{equation*}
The model parameters $a = 0.77$ and $b = 0.63$ were used \cite{lodge2000parametric}. MFR was calculated as the ratio between the stress MBF and rest MBF. The IDIF was obtained using manually drawn volumes of interest (VOIs) in the left ventricle. $K_1$, MBF, and $k_2$ were calculated as the mean values within the manually drawn VOIs of the myocardium region. 

\begin{figure*}[htb!]
    \centering
    \includegraphics[width=\linewidth]{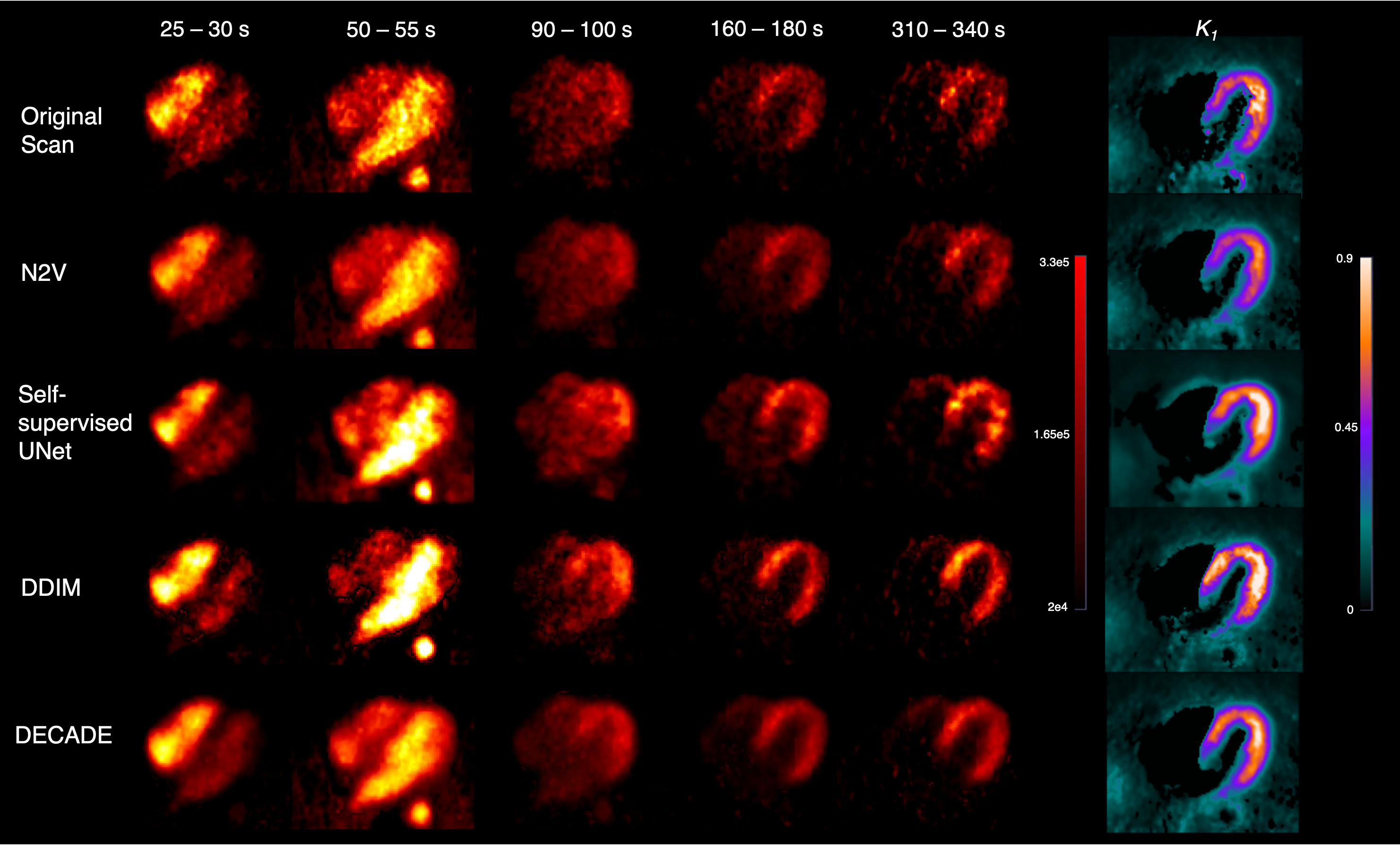}
    \caption{Example dynamic frames and $K_1$ images of a rest scan from the Vision 450 dataset. The radiology report indicates the patient has no myocardial defect. DECADE achieved superior denoising performance across all dynamic frames. The denoised frames and $K_1$ map produced by DECADE exhibit the lowest noise levels and the most similar tracer distribution to the original dynamic frames. }
    \label{comparison_vision}
\end{figure*}

\begin{table*}[htb!]
    \centering
    \footnotesize
    \caption{Mean K1, MBF, and MFR values for 50 subjects with rest and stress scans acquired on Siemens Vision 450. }
    \label{vision450}
     \begin{adjustbox}{width=0.9\textwidth}
        \begin{tabularx}{0.9\textwidth}{cYY|YY|Y}
        \hline
        \multirow{2}{*}{ } & \multicolumn{2}{c|}{Rest} & \multicolumn{2}{c|}{Stress} & \multirow{2}{*}{MFR} \\ \cline{2-5}
        & $K_1$ (ml/min/g) & MBF (ml/min/g) & $K_1$ (ml/min/g) & MBF (ml/min/g) & \\ \hline
        Noisy Frames & 0.61 ±  0.17 & 1.11 ± 0.53 & 0.84 ± 0.23 & 1.92 ± 0.79 & 1.86 ± 0.74 \\ 
        N2V & 0.56 ± 0.14 & 0.95 ± 0.44 & 0.76 ± 0.20 & 1.64 ± 0.65 & 1.82 ± 0.66 \\ 
        Self-supervised UNet & 0.70 ± 0.19 & 1.41 ± 0.65 & 1.01 ± 0.29 & 2.59 ± 1.05 & 1.96 ± 0.79 \\ 
        DDIM & 0.78 ± 0.25 & 1.71 ± 0.93 & 1.22 ± 0.49 & 3.42 ± 1.50 & 2.22 ± 1.13 \\ 
        DECADE & 0.65 ± 0.17 & 1.21 ± 0.56 & 0.89 ± 0.23 & 2.11 ± 0.82 & 1.89 ± 0.74 \\ \hline
           
        \end{tabularx}
    \end{adjustbox}
    
\end{table*}
 
\subsection{Evaluation Metrics and Baseline Comparison}
To evaluate the proposed method, the image quality of the original dynamic scans and the denoised scans were visually compared for the Vision 450 data. The quantitative accuracy of MBF and MFR values was also assessed. Additionally, the method was validated using the high-sensitivity Quadra scanner dataset, where low-count and full-count pairs were created. Full-count images were used as the reference to compare with the denoised results of low-count images. For the Quadra scanner dataset, we assessed the image quality and quantification accuracy of denoised dynamic frames and parametric images derived from the denoised frames. Image quality metrics included Peak Signal-to-Noise Ratio (PSNR), Structural Similarity Index (SSIM), and Normalized Mean Square Error (NMSE) between the denoised outputs and full-count images. For parametric images, we compared the percentage difference between the MBF values of denoised and full-count images:

\begin{equation*}
    \mathrm{Percentage\ Error} = \frac{|MBF_{pred} -MBF_{true}|}{MBF_{true}} * 100\%
\end{equation*}
The percentage difference of MFR values was calculated similarly. 

Two U-Net based models and a diffusion-based model are compared with the proposed method. Noise2Void (N2V) is a self-supervised denoising method that trains a neural network without the need for clean target images. Instead of learning from paired noisy-clean image datasets, N2V utilizes only noisy images by masking out certain pixels and predicting their values based on the surrounding context \cite{krull2019noise2void}. Self-supervised U-Net is a uniquely designed noise-aware method for dynamic PET denoising \cite{xie2025noise}, which is built on N2V and encoded noise level information of dynamic frames into the U-Net, achieving state-of-the-art performance for \textsuperscript{82}Rb dynamic cardiac PET denoising. The Denoising Diffusion Implicit Model (DDIM) generalizes the DDPM to a non-Markovian, deterministic path for sampling that enables faster and more consistent sampling \cite{song2020denoising}. For DDIM training, since it is impossible to obtain clean-noisy image pairs for individual dynamic frames, we downsampled the clean mean static images into 5\%, 10\%, 25\%, and 50\% low-count images to simulate different noise levels across dynamic frames and trained a supervised conditional diffusion model with paired low-count and full-count mean static images. 

\subsection{Implementation Details}

We implemented our method in PyTorch and conducted experiments using an NVIDIA H100 GPU with 80 GB of memory. During Phase A, the unconditional diffusion model was optimized using the Adam optimizer with a learning rate of 1e-4, a batch size of 10, and a cosine scheduler. The model was trained for 300,000 steps, and the best model was selected based on the validation dataset. In Phase B, when training the ControlNet, the pre-trained diffusion models were frozen, and the trainable copy blocks were optimized using the Adam optimizer with a learning rate of 5e-5 and a batch size of 10. The ControlNet was trained for 100,000 steps. The sampling process started from Gaussian noise and recovered images in $T=1000$ steps, with a constraint weight of 500 used for the regularization term incorporating the noisy frame.

\begin{figure*}[htb!]
    \centering
    \includegraphics[width=\linewidth]{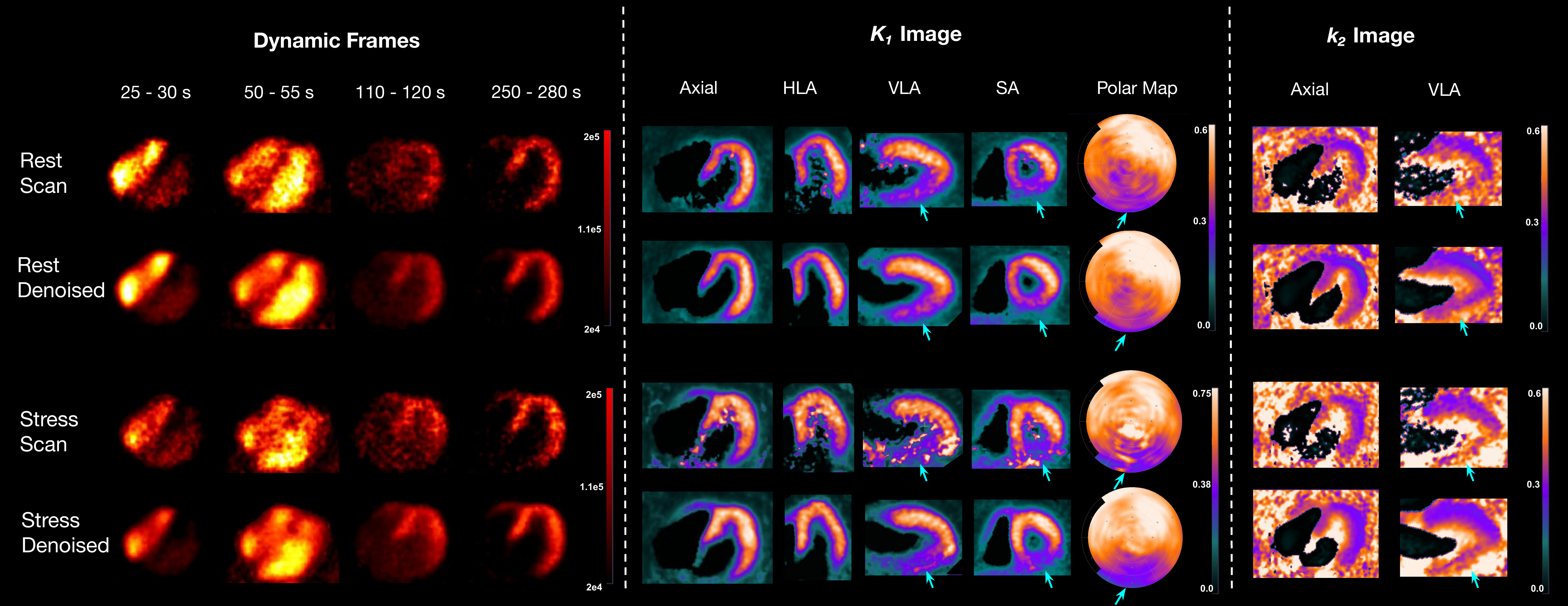}
    \caption{An example subject from the Vision 450 dataset with a confirmed myocardial defect. The clinical report indicates a partially reversible perfusion defect in the basal to apical inferior, basal inferolateral, and basal inferoseptal walls. The catheteriztion study confirms the appearance of dominant LCx with severe calcified stenosis. Dynamic frames, $K_1$ images, $k_2$ images, and polar maps before and after denoising are shown. $K_1$ images are displayed in axial view, horizontal long-axis view (HLA), vertical long-axis view (VLA), and short-axis view (SA). DECADE achieves superior denoising performance across all dynamic frames over time, producing $K_1$ maps with reduced noise. Notably, the stress polar map's noise artifact near the apex region disappears after denoising, and the boundary of the myocardium becomes more apparent. Importantly, the myocardial defect, pointed out by a blue arrow in the VLA and SA views of parametric images and polar map, remains visible and even easier to detect after denoising due to improved parametric image quality. The defect in the $k_2$ image is also indicated by higher uptake in the defect region of the VLA and SA view.}
    \label{abnormal}
\end{figure*}

\begin{figure*}[htb!]
    \centering
    \includegraphics[width=\linewidth]{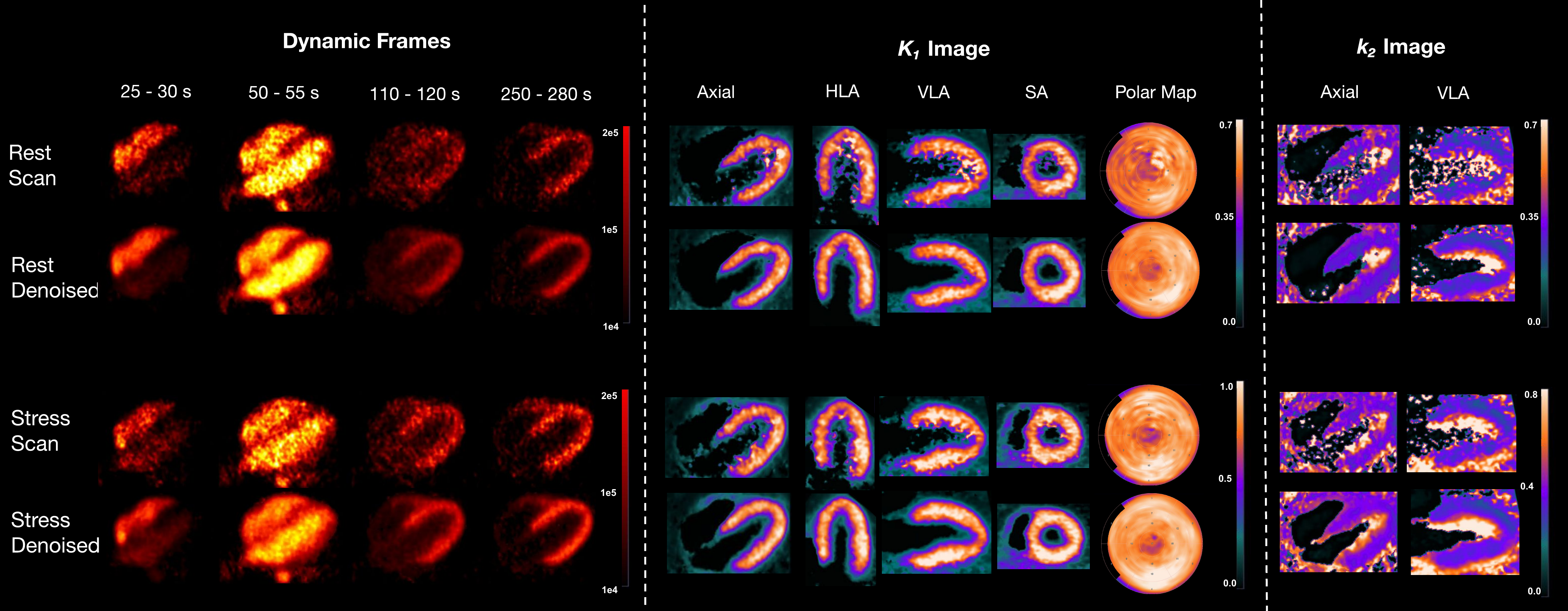}
    \caption{A normal subject from the Vision 450 dataset. The radiology report indicates this patient has no myocardial defects. DECADE also produces visually appealing denoised results for this patient on dynamic frames. The denoised images exhibit a more uniform distribution and significantly lower noise levels. The contour of the myocardium is better depicted in all views of the $K_1$ and $k_2$ images. }
    \label{normal}
\end{figure*}

\begin{figure*}[htb!]
    \centering
    \includegraphics[width=\linewidth]{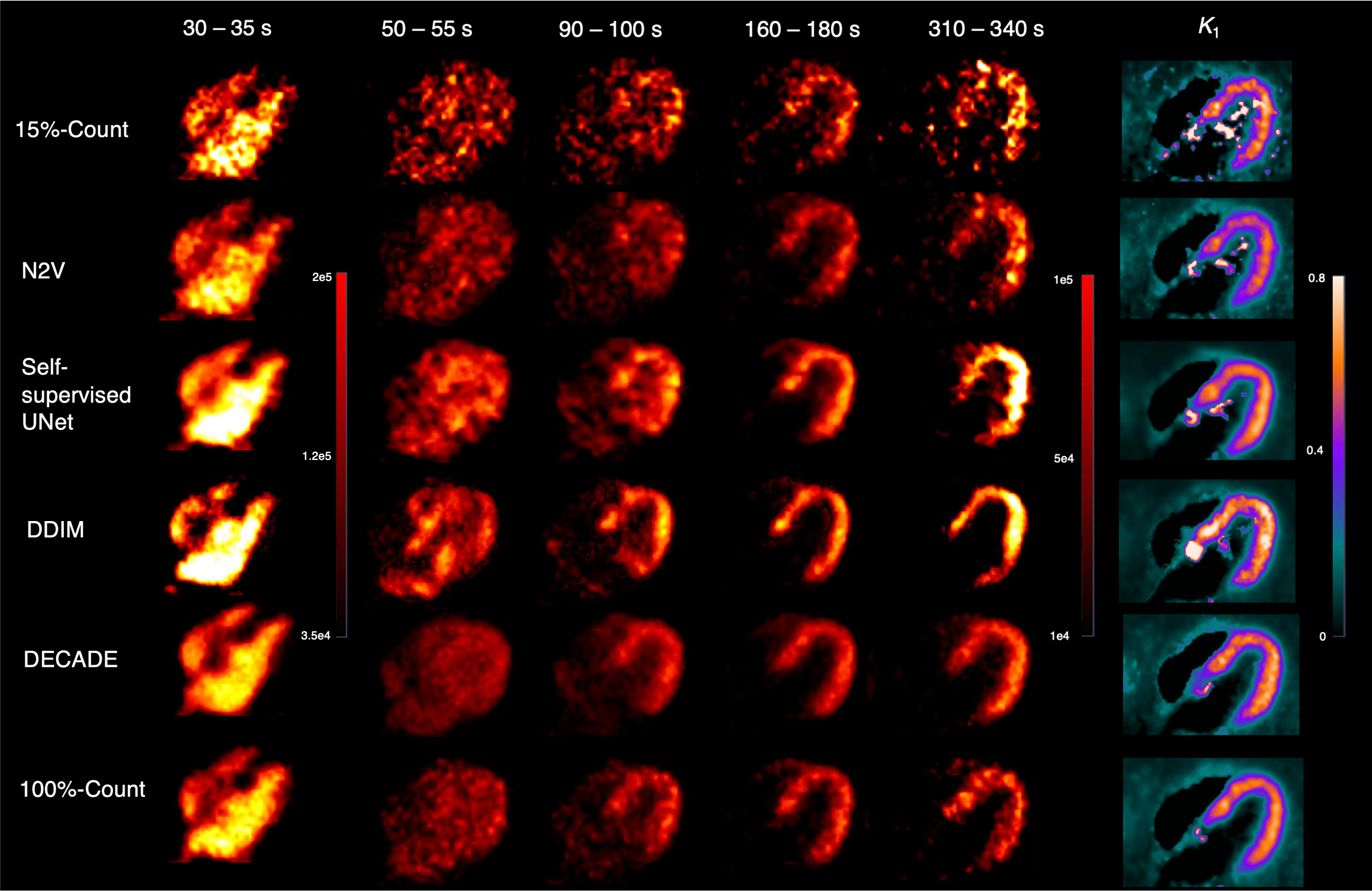}
    \caption{Comparison results on an example rest scan from the Quadra Bern dataset. 15\% dynamic frames are used as noisy inputs, and 100\% scans are used as ground truth to compare with denoised results. DECADE successfully generalizes to the downsampled Quadra Bern dataset without fine-tuning or re-training. The denoised images from DECADE preserve structural details of anatomical regions and ensure quantitative accuracy. In contrast, underestimation occurs with N2V, and overestimation occurs with the self-supervised UNet and DDIM for dynamic frames.  }
    \label{quadra_comp}
\end{figure*}

\section{Results}

\subsection{Denoised Results on Vision 450 Dataset}

The visual comparison between the original dynamic frames and the denoised frames using different denoising approaches is shown in Fig. \ref{comparison_vision}. Five sample dynamic frames from early to late stages are depicted. It is evident that the image quality of the denoised images is substantially improved with DECADE compared to other denoising approaches, while preserving structural details. DECADE consistently achieves superior denoising performance across all dynamic frames. For dynamic PET, early frames are less noisy, while late frames exhibit substantial noise due to the decay of \textsuperscript{82}Rb reducing tracer concentration in the cardiac region. DECADE effectively maintains a clear shape and boundary of the myocardium, even in the late frames.

Compared with N2V and the self-supervised U-Net, DECADE significantly reduces noise levels, resulting in higher-quality images with clearer visualization of anatomical structures through the iterative recovery steps of diffusion models. Although DDIM successfully provides denoised images with a correct shape, the overall images are substantially overestimated compared to DECADE, which incorporates noisy frames as constraints in the sampling process to ensure quantification accuracy. The corresponding $K_1$ parametric images for each method are also included. Similarly, the $K_1$ images from DECADE show qualitatively accurate results, with the most uniform distribution throughout the myocardium for this normal patient, particularly in the apex region, which exhibits slightly higher uptake. The comparison of the image-derived input function from the left ventricle ROI is shown in the Supplement. Table \ref{vision450} summarizes the overall $K_1$, MBF, and MFR values for different methods. Table \ref{vision450} summarizes the overall $K_1$, MBF, and MFR values for different methods. DECADE achieves effective denoising without compromising quantification accuracy, ensuring that the regional $K_1$, MBF, and MFR values of the myocardial region of intrest (ROI) on denoised images remain consistent with those from the original dynamic frames. 

% Table \ref{vision450} summarizes overall $K_1$, MBF, and MFR values for different methods. N2V underestimates MBF and MFR values in the myocardium, while the self-supervised UNet and DDIM overestimate these values. By contrast, DECADE achieves accurate 
% $K_1$, MBF, and MFR values compared to the original dynamic frames, while also significantly improving the image quality of the denoised frames and parametric images.

Fig. \ref{abnormal} and \ref{normal} display denoised results for a patient with myocardial defects and a normal patient, respectively. The clinical report for the patient in Fig. \ref{abnormal} indicates a partially reversible perfusion defect in the basal to apical inferior, basal inferolateral, and basal inferoseptal walls. The appearance of the dominant LCx with high grade, severe calcified stenosis is confirmed by a cardiac catheterization study. The denoised images show a significant reduction in noise, particularly in late frames (250-280 s), where noticeable noise is visible in the septal wall for rest scans and near the apical region for stress scans. DECADE successfully enhances the visualization of tracer distribution by significantly reducing the noise level and improving overall image quality. Noisy dynamic frames lead to noisy parametric images; for example, the $K_1$ image of stress scans shows high-noise regions in the apex, in which noise artifacts are significantly reduced after denoising with DECADE, preserving quantification across the myocardium. The denoised $K_1$ images also have clearer boundaries and cleaner visualization in the left ventricle. Most importantly, the myocardial defect, indicated by the blue arrow in the VLA view and polar map, remains visible after denoising, demonstrating DECADE's ability to reduce noise while maintaining the detectability of myocardial defects. The $k_2$ images also exhibit a clearer myocardial contour and a substantial reduction in overall noise levels after denoising.

Fig. \ref{normal} presents the denoising results for a normal patient without myocardial defects. DECADE improves the image quality of dynamic frames and parametric images, reduces noise levels, and maintains the structure of the myocardium for all denoised results. These observations demonstrate the effectiveness of our denoising approach for both normal and abnormal patients with myocardial defects. To quantify the improvement in image quality of parametric images, the noise level of $K_1$ images was evaluated using normalized standard deviation (NSTD) in the myocardial ROI on normal subjects, whose scans are expected to have uniform distribution. Among 30 normal subjects out of 50 testing subjects, the NSTD improves from 0.147 $\pm$ 0.090 to 0.117 $\pm$ 0.036 for rest scans and from 0.130 $\pm$ 0.042 to 0.110 $\pm$ 0.027 for stress scans before and after denoising, respectively. The myocardium-to-blood ratio of the static frame increases from 2.175 $\pm$ 0.406 to 2.401 $\pm$ 0.452 for rest scans and from 2.632 $\pm$ 0.498 to 2.887 $\pm$ 0.565 before and after denoising, respectively, across all 50 testing subjects, indicating better image contrast.

\begin{table}[t]
    \centering
    \footnotesize
    \caption{Comparison across different denoising methods applied to 15\% count-level images, with 100\% count images serving as the reference, from 5 subjects scanned using Siemens Biograph Vision Quadra scanners (250 images in total) }
    \label{time}
    \begin{tabular}{cccc}
    \hline
     & SSIM $\uparrow$ & PSNR $\uparrow$ & NMSE $\downarrow$ \\ \hline
    Noisy Frames & 0.59 ± 0.12 & 26.72 ± 4.03 & 0.47 ± 0.16 \\ 
    N2V & 0.66 ± 0.11 & 27.82 ± 3.09 & 0.40 ± 0.10 \\ 
    Self-supervised UNet & 0.68 ± 0.07 & 26.84 ± 2.23 & 0.40 ± 0.10 \\ 
    DDIM & 0.58 ± 0.07 & 24.02 ± 2.94 & 0.61 ± 0.11 \\ 
    \textbf{DECADE} & \textbf{0.73 ± 0.10} & \textbf{30.64 ± 3.40} & \textbf{0.29 ± 0.08} \\ \hline
    \end{tabular}
    \label{quadra}
\end{table}

\subsection{Validation on the LAFOV Quadra Dataset}
The proposed method was also validated on the LAFOV Quadra dataset from Bern, where the low-count 15\% frames were used as noisy input and the denoised results were compared with full-count scans. Fig. \ref{quadra_comp} shows the comparison of denoised frames and $K_1$ images using different denoising approaches. Among all denoising methods, DECADE most closely resembles the full-count scan and achieves the best image quality. DECADE shows clear anatomical structures after denoising and the most uniform tracer uptake. While other methods achieved reasonable denoising performance in early frames, artifacts appeared in the denoised images for late frames as noise levels increased. For example, N2V maintains consistent shape with the original scans, but the noise level remains high over time. The self-supervised UNet and DDIM generate incorrect high noise in the myocardium region for frames from 160 to 180 seconds and 310 to 340 seconds. Anatomical details of the cardiac region are also distorted for frames from 50 to 55 seconds and 90 to 100 seconds. By contrast, DECADE shows clear anatomy of the heart, including the left ventricle and right ventricle in the early phase and the myocardium in the late phase. The comparison of the image-derived input function from the left ventricle ROI for this case is shown in the Supplement.

The quantitative accuracy was also validated on five subjects with rest and stress scans from the Bern Quadra dataset, shown in Table \ref{quadra}, including 25 dynamic frames for each scan and 250 individual frames in total. Full-count frames were used as reference images to compare with denoised images. DECADE achieved the highest SSIM of 0.73, PSNR of 30.64, and the lowest NMSE of 0.29, averaging across all frames. Compared with the noisy frame, DECADE improved SSIM by 23.7\%, PSNR by 14.7\%, and NMSE by 38.3\%. It also surpassed the second-best denoising method in SSIM by 7.4\%, PSNR by 10.1\%, and NMSE by 27.5\%, indicating its superior denoising performance. A paired t-test confirms that the improvement in SSIM, PSNR, and NMSE from all other methods and the original dynamic frames is statistically significant ($p < 0.0001$).

DECADE also achieved superior qualitative and quantification results for tracer kinetic modeling and parametric images as shown in Fig. \ref{quadra_comp} and Table \ref{quadra_k1}. $K_1$ images derived from denoised frames by DECADE present the lowest noise and the most consistent distribution compared to full-count images in the myocardium, with the highest image quality. The quantification accuracy of $K_1$, MBF, and MFR was further validated by calculating their mean values in the manually drawn myocardium ROI regions. Among all methods, DECADE achieved the lowest average MBF error for both rest and stress scans. It reduced the MBF error by 2.34\%  from 13.95\% to 11.61\% for the rest scan and by 2.55\% from 11.04\% to 8.49\% for the stress scan compared with the second-best method, self-supervised UNet. MFR values from DECADE were also closest to full-count frames, with a percentage error of 10.86\%, surpassing the second-best method, N2V, by 2.15\%, indicating its ability to accurately denoise images and produce quantitatively accurate parametric images for both rest scans and stress scans. 

\begin{table*}[htb!]
    \centering
    \footnotesize
    \caption{Mean K1, MBF, and MFR values for 5 subjects with rest and stress scans acquired on Siemens Quadra scanner. 100\% count is used as the reference image.}
    
     \begin{adjustbox}{width=\textwidth}
        \begin{tabularx}{\textwidth}{cccY|ccY|cY}
        \hline
        \multirow{4}{*}{ } & \multicolumn{3}{c|}{Rest} & \multicolumn{3}{c|}{Stress} & \multirow{3}{*}{MFR} & \multirow{3}{*}{MFR Error(\%)$\downarrow$ }\\ \cline{2-7}
        & \multirow{1}{*}{$K_1$} & \multirow{1}{*}{MBF} & \multirow{2}{*}{MBF Error(\%)$\downarrow$} & \multirow{1}{*}{$K_1$} & \multirow{1}{*}{MBF} & \multirow{2}{*}{MBF Error(\%)$\downarrow$} & & \\ 
        % & & & & & & & & \\
        & (ml/min/g) & (ml/min/g) & & (ml/min/g) & (ml/min/g) & & & \\ \hline
        % 15\% Frames & 0.39 ± 0.17 & 0.56 ± 0.37 & 10.30 ± 9.27 & 0.63 ± 0.25 & 1.23 ± 0.82 & 5.64 ± 6.48 & 2.21 ± 0.66 & 12.53 ± 5.46 \\
        N2V & 0.36 ± 0.15 & 0.48 ± 0.31 & 17.42 ± 11.22 & 0.57 ± 0.21 & 1.01 ± 0.64 & 16.63 ± 7.78 & 2.10 ± 0.52 & 12.91 ± 5.95 \\ 
        Self-supervised UNet & 0.38 ± 0.15 & 0.53 ± 0.31 & 13.95 ± 9.62 & 0.60 ± 0.25 & 1.13 ± 0.78 & 11.04 ± 8.61 & 2.09 ± 0.52 & 17.67 ± 7.58 \\
        DDIM & 0.51 ± 0.15 & 0.81 ± 0.38 & 54.90 ± 70.48 & 0.75 ± 0.29 & 1.62 ± 0.99 & 31.83 ± 20.07 & 1.94 ± 0.60 & 27.76 ± 9.34 \\
        \textbf{DECADE} & \textbf{0.42 ± 0.17} & \textbf{0.62 ± 0.39} & \textbf{11.61 ± 7.95} & \textbf{0.65 ± 0.25} & \textbf{1.26 ± 0.83} & \textbf{8.49 ± 5.50} & \textbf{2.14 ± 0.55} & \textbf{10.86 ± 7.66} \\
        100\% Frames & 0.41 ± 0.16 & 0.58 ± 0.38 &  & 0.63 ± 0.22 & 1.19 ± 0.70 &  & 2.28 ± 0.86 & \\ \hline
           
        \end{tabularx}
    \end{adjustbox}
    \label{quadra_k1}
    
\end{table*}

\begin{figure*}[htb!]
    \centering
    \includegraphics[width=\linewidth]{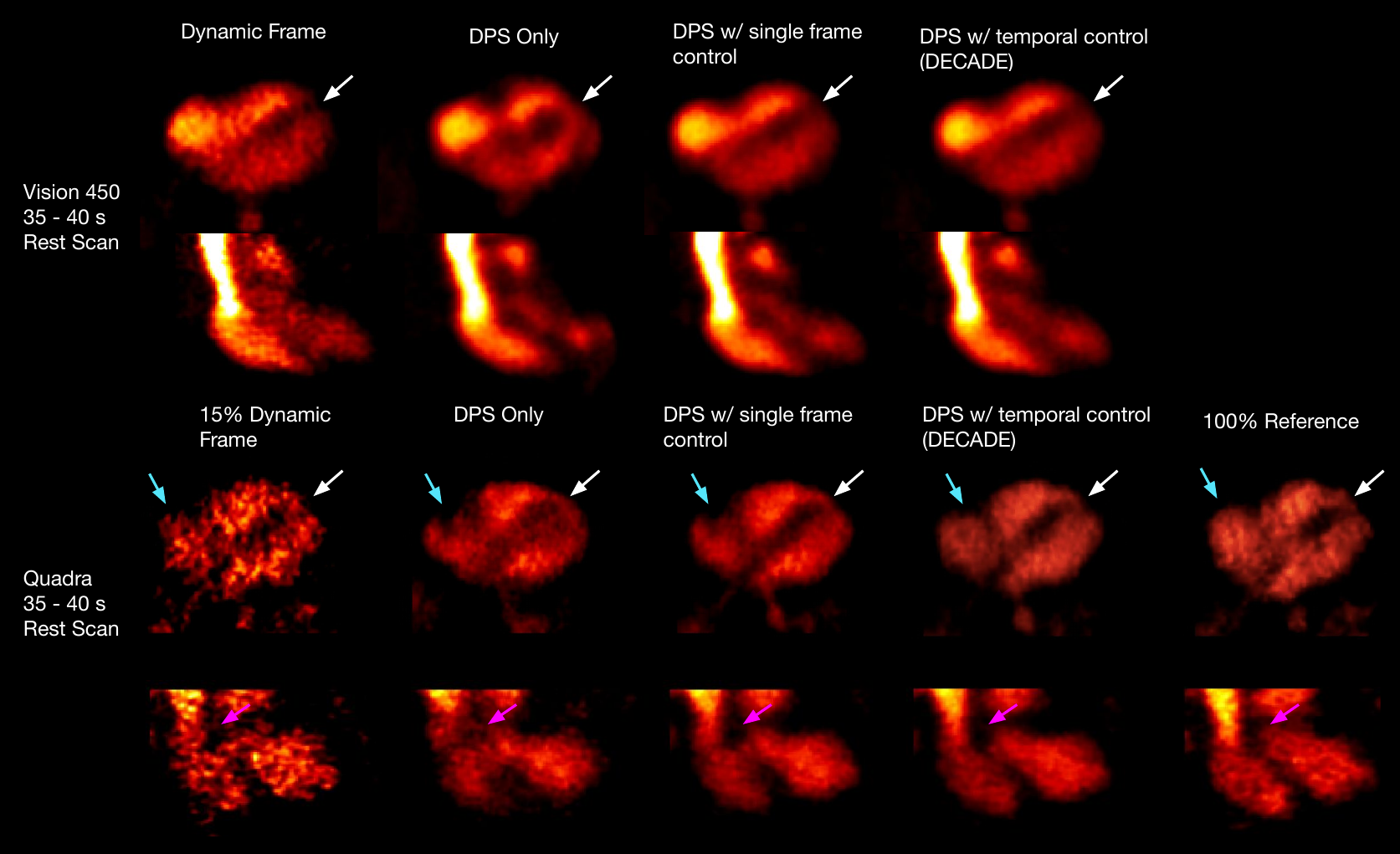}
    \caption{The ablation study on the effect of image guidance and temporal consistency using both the Vision 450 dataset and the Quadra dataset. The early frame of a rest scan is shown. Adding image guidance using DPS ensures quantitative accuracy similar to the original dynamic frames, as indicated by the "DPS Only" column. Refining the pre-trained diffusion model using noisy frames through ControlNet further enhances qualitative accuracy and helps preserve structural details in anatomical regions with weak signals. The "DPS w/Single Frame Control" uses the current Gaussian-smoothed frame as control, while DECADE uses three consecutive frames to add temporal consistency. Using multiple frames as conditional input further increases the model's performance in more challenging regions with relatively weak signals, such as the cavity between the left ventricle and right ventricle, or the edges of anatomical regions, resulting in the best denoised images. Colored arrows indicate artifacts that appear where temporal consistency and image guidance are not added. The white arrow points to the connecting artifact in the septal wall, the cyan arrow points to the edge of the right atrium wall, and the purple arrow points to the boundary between the vena cava and the right atrium.}
    \label{ablation}
\end{figure*}

\subsection{Ablation Studies}
Fig. \ref{ablation} displays the qualitative results from the ablation study on the proposed components of image guidance in the sampling stage and temporal consistency for refining the diffusion model. One example frame from the Vision 450 dataset and another from the Quadra scanner are shown. 

Even though the diffusion model is pre-trained using only 2-7 minute mean static images, DPS allows the model to utilize the prior information learned from the training data and generalize well to the less noisy early frames, as shown in the denoised image with only DPS. The image guidance ensures that the pre-trained diffusion models can iteratively recover denoised images from the Gaussian distribution consistent with the noisy frames. Notably, DPS alone achieves better performance than the second-best method in Table \ref{quadra} for PSNR and NMSE, indicating the superior performance of our unsupervised diffusion method. However, since the pre-trained diffusion models did not incorporate noisy frames during training, the denoised results may be sub-optimal due to the discrepancy between the training data distribution and dynamic data distribution, and hallucinations could occur in pixels with low uptake, leading to lower SSIM.

For example, in the early frames (35-40 s) shown in Fig. \ref{ablation} for both the Vision 450 and Quadra data, a connecting artifact in the septal wall (indicated by a white arrow) is observed in the denoised image using only DPS. After incorporating Gaussian-smoothed noisy frames into the pre-trained diffusion models, these artifacts are alleviated, and the cavity between the left and right ventricles becomes more visible. Temporal consistency further refines the anatomical details of the denoised images. In the Vision 450 dataset, adding temporal consistency produces a more contiguous anatomical structure in the left ventricle. In the Quadra dataset, only DECADE recovers the oval shape near the right atrium wall in the axial view (indicated by the cyan arrow), while part of the right atrium is missing in the other two methods. In the coronal view, DECADE also produces a more homogeneous structure in the right atrium and a clearer boundary of the vena cava and left atrium (indicated by the purple arrow).

The effectiveness of temporal consistency and image guidance is confirmed by quantitative metrics shown in Table \ref{ablation_quant}, where the addition of image guidance and temporal consistency progressively improves SSIM, PSNR, and NMSE. DPS improves image quality and quantification accuracy from noisy dynamic frames, and incorporating noisy frames into pre-trained diffusion models significantly enhances image quality, as indicated by DPS with single frame control and DECADE. The notable improvement in SSIM is also observed. With consecutive frames as temporal control in diffusion training and sampling, DECADE achieves the best denoising performance across all evaluation metrics.

\begin{table}[h]
    \centering
    \footnotesize
    \caption{Ablation studies on 5 subjects scanned using Siemens Biograph Vision Quadra scanners (250 images in total), 15\% frames are used as input images and full-count images are used as the reference.}
    \label{time}
    \begin{tabular}{cccc}
    \hline
     & SSIM $\uparrow$ & PSNR $\uparrow$ & NMSE $\downarrow$ \\ \hline
     15\% Dynamic Frames & 0.587 & 26.724 & 0.469 \\
     DPS Only & 0.603 & 28.093 & 0.388 \\ 
     DPS w/ single frame control & 0.717 & 30.063 & 0.315 \\ 
    \textbf{DECADE} & \textbf{0.733} & \textbf{30.641} & \textbf{0.291} \\ \hline
    \end{tabular}
    \label{ablation_quant}
\end{table}

\section{Discussion}
In this study, we propose an unsupervised diffusion model framework, DECADE, that integrates image guidance and temporal consistency in its diffusion training and sampling stages. This framework generalizes to all dynamic frames with distinctive tracer distributions and achieves both qualitatively and quantitatively accurate denoising for all frames. Our proposed method addresses the challenge of lacking paired high-resolution and noisy image training data for \textsuperscript{82}Rb cardiac dynamic PET. The 3D diffusion model is pre-trained on 2-7 minute mean static PET images to learn prior information from pseudo-clean images. Gaussian-smoothed consecutive noisy frames are then incorporated via ControlNet to refine the pre-trained diffusion model for denoising dynamic frames.

Additionally, we addressed the challenge of rapid tracer distribution changes across dynamic frames in our diffusion sampling strategy, generalizing our approach to different frames. This was achieved by incorporating individual noisy frames as image guidance through a regularization term between the denoised images at each time step and the noisy frame in the DPS sampling process, ensuring quantification accuracy during denoising. In contrast, the conditional DDIM, trained on paired low-count and full-count mean static images, failed to preserve quantification accuracy and overestimated denoising on dynamic frames with zero-shot inference, likely due to domain differences between mean static images and individual dynamic frames. DECADE also demonstrates better generalization on the Quadra scanner dataset compared with other methods. More distortions of anatomical structures and incorrect predictions occur more frequently in the denoised images of other methods in regions with high noise, especially from mid to late frames when counts are lower and dynamic frames are noisier.

A unique characteristic of dynamic PET is the temporal information throughout the acquisition process. Our method integrates this temporal information by using neighboring frames to leverage additional context, where consecutive frames are used as conditional input to refine the pre-trained diffusion models via ControlNet, which is then used in the sampling stage. We hypothesized that tracer distribution continuity between neighboring frames can prevent hallucinations in areas with weak signals, such as the cavity between the left and right ventricles and the edges of anatomical structures. Temporal consistency through ControlNet indeed helps preserve anatomical details and reduce artifacts, as demonstrated in the ablation study.

We acknowledge that limitations still exist in the current framework. First, only three consecutive frames are currently used to ensure temporal consistency. However, utilizing temporal information from the entire acquisition time could enhance model learning on 4D data and produce more consistent results across all frames for the same patient. For example, embedding the frame index and time-activity curve for different ROIs, including the left ventricle, right ventricle, and myocardium, could provide additional context \cite{guo2024tai}. Second, incorporating more anatomical information could ensure image domain consistency. CT scans available for each study could provide accurate anatomical contours to improve denoising performance, similar to methods that use brain MRI for brain PET denoising \cite{onishi2021anatomical}. Segmentation masks of the heart could also be added to the diffusion model for guided denoising \cite{konz2024anatomically}. Lastly, in the current framework, the pre-training stage of the diffusion model only includes 2-7 minute mean static PET images because early frames change rapidly and clean images for early frames do not exist. However, the low-quality issues of dynamic PET mainly occur in the late frames, whereas the counts in the early stage are still relatively high. As a result, it is more challenging for the model to achieve optimal denoising performance on late frames. Early frame information is also included in the second training stage when refining the pre-trained model with ControlNet. However, if a large number of high-resolution images were available for all dynamic frames, such as Quadra scans, these could be used to pre-train the diffusion model, enabling it to learn prior information on different tracer distributions for all dynamic frames. Additionally, it is possible to create a labeled dataset with paired full-count and low-count frames for training a supervised diffusion model, then add temporal consistency through ControlNet to further enhance performance.

\section{Conclusion}
In our study, we introduce DECADE, a temporally-consistent unsupervised diffusion model for \textsuperscript{82}Rb dynamic cardiac PET denoising. By leveraging temporal information from neighboring frames and image guidance from individual noisy frames, DECADE does not require paired data in its training and sampling processes. DECADE enables unsupervised learning by initially pre-training a 3D diffusion model to learn prior information from 2-7 minute mean static images, followed by refining this pre-trained model with ControlNet using temporal context from consecutive noisy dynamic frames. Finally, the proposed diffusion sampling strategy integrates image guidance and temporal information to enhance both the measurement fidelity and perceptual quality of the denoised outputs. The proposed method has been validated on both the Vision 450 dataset and the Quadra dataset, achieving superior denoising performance across all dynamic frames and generalizing well to different datasets. The denoised images generated by DECADE show significant improvements in image quality and maintain clear anatomical structures. The parametric images also demonstrate accurate quantification and diagnostic accuracy in both patients with and without myocardial defects. 

% -------------------------------------------------------------------
\section*{Acknowledgments}
This work was supported by the National Institutes of Health (NIH) grant R01HL169868.

% -------------------------------------------------------------------
\section*{Declaration of Competing Interest}
The authors declare that they have no known competing financial interests or personal relationships that could have appeared to influence the work reported in this paper.

% {\appendix[Proof of the Zonklar Equations]}

\bibliography{reference}
\bibliographystyle{IEEEtran}

\end{document}